\documentclass[letterpaper, 10 pt, conference]{ieeeconf}  

\IEEEoverridecommandlockouts                              

\usepackage{cite}
\usepackage{amsmath,amssymb,amsfonts}
\usepackage{algorithmic}
\usepackage{graphicx}
\usepackage{textcomp}
\usepackage{xcolor}
\usepackage{hyperref}
\usepackage{booktabs}
\usepackage{multirow}
\usepackage{subcaption}
\def\BibTeX{{\rm B\kern-.05em{\sc i\kern-.025em b}\kern-.08em
    T\kern-.1667em\lower.7ex\hbox{E}\kern-.125emX}}
\usepackage[capitalize]{cleveref}
\crefname{section}{Sec.}{Secs.}
\Crefname{section}{Section}{Sections}
\Crefname{table}{Table}{Tables}
\crefname{table}{Tab.}{Tabs.}

\begin{document}

\title{Interpretable End-to-End Driving Model for Implicit Scene Understanding
}

\author{Yiyang Sun,  Xiaonian Wang, Yangyang Zhang, Jiagui Tang, Xiaqiang Tang, and Jing Yao$^{\dag}$%
\thanks{Y.Sun, X.Wang, Y.Zhang, J.Tang, X.Tang, and J.Yao are with the College of Electronical and Information Engineering, Tongji University, China. E-mail: \{2130734, dawnyear, yangyangzhang, 2130701, 2130705, yaojing\}@tongji.edu.cn}%
\thanks{$^{\dag}$ Indicates the corresponding author.}
}

\maketitle

\begin{abstract}
Driving scene understanding is to obtain comprehensive scene information through the sensor data and provide a basis for downstream tasks, which is indispensable for the safety of self-driving vehicles. Specific perception tasks, such as object detection and scene graph generation, are commonly used. However, the results of these tasks are only equivalent to the characterization of sampling from high-dimensional scene features, which are not sufficient to represent the scenario. In addition, the goal of perception tasks is inconsistent with human driving that just focuses on what may affect the ego-trajectory. Therefore, we propose an end-to-end Interpretable Implicit Driving Scene Understanding (II-DSU) model to extract implicit high-dimensional scene features as scene understanding results guided by a planning module and to validate the plausibility of scene understanding using auxiliary perception tasks for visualization. Experimental results on CARLA benchmarks show that our approach achieves the new state-of-the-art and is able to obtain scene features that embody richer scene information relevant to driving, enabling superior performance of the downstream planning.
\end{abstract}


\section{Introduction}
\begin{figure}[t]
        \centering
        \includegraphics[clip, width=8.2cm]{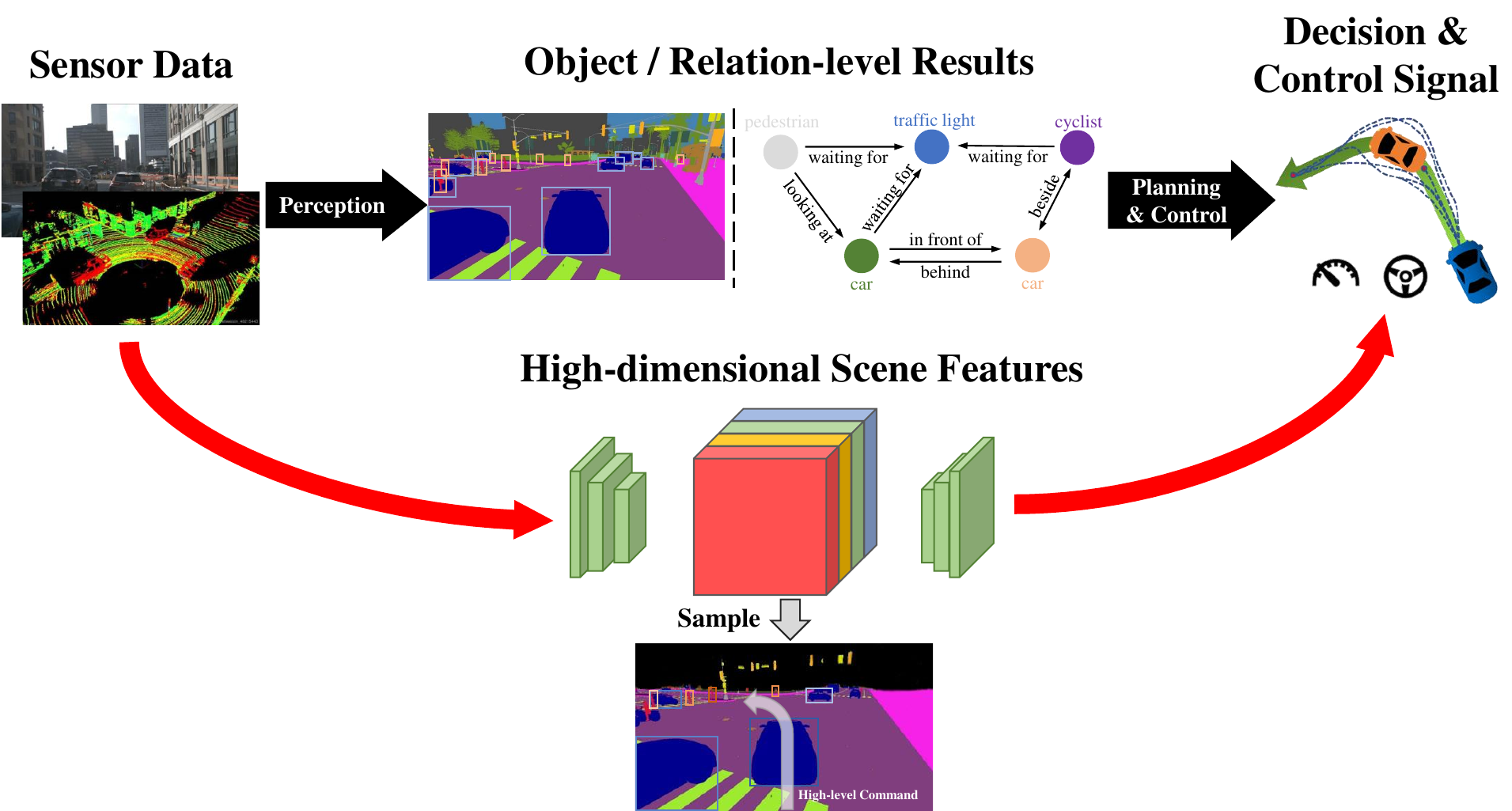}
        \caption{\textbf{Two architectures of the autonomous driving system.} The upper line (Black) shows the sequential architecture, which uses the explicit perception task as scene understanding. The lower line ({\color {red} {red}}) is the end-to-end model whose scene feature is high-dimensional. Compare the difference between the two, end-to-end model only pays attention to what will influence the planning trajectory, which is consistent with human habits.
        }
        \label{fig:pipeline}
        \vspace{-3mm}
    \end{figure}

    The current typical system for autonomous driving is mainly a sequential architecture~\cite{gog2021pylot} consisting of three modules as illustrated in the above line of ~\cref{fig:pipeline}. It divides the whole autonomous driving task into three cascade parts. First, the perception part transforms the sensor data into a comprehensive understanding of the scenario, including the localization of the ego vehicle using a high-definition (HD) map and the information on dynamic and static obstacles. Next, the decision-making and planning part is used to generate a reasonable future trajectory of the ego vehicle based on the perception results. Finally, the control part issues commands for controlling the vehicle and tracking the planned trajectory.
    
    However, for the driving scene understanding task, using a sequential architecture system has several disadvantages. {\bf Low-dimension understanding.} The common perception tasks used in the sequential architecture system are object detection~\cite{lang2019pointpillars}, trajectory prediction~\cite{gao2020vectornet}, semantic segmentation~\cite{chen2022self}, etc. Nevertheless, the output of these tasks is basically object-level and only corresponds to an explicit representation of some dimensions sampled in the high-dimensional feature space represented by the whole scenario. Although there are up-and-coming object-relation-level tasks such as scene graph generation~\cite{tian2021rsg, malawade2022spatiotemporal} used for driving scene understanding, they don't solve this problem. In other words, a comprehensive understanding of the driving scene cannot be accomplished by using several explicit perception tasks. {\bf Mis-aligned goal.} The objects of the perception tasks are not aligned with the overall goal of self-driving~\cite{wei2021perceive}. Take object detection as an example, the ultimate goal of self-driving is to drive safely to the destination, which means that only the objects which will affect the driving trajectory need to be focused on and detected. But the goal of the object detection task is to detect all objects in the whole scene to achieve improved metrics. This leads to a large amount of computation and model capacity being wasted on identifying difficult instances that are unhelpful to the driving process. {\bf Complicated interface.} Since the whole autonomous driving task is separated into three sub-tasks, the output format of the perception part should be defined explicitly in advance as well as the interface of the downstream planning part. 

    To address these issues, we propose Interpretable Implicit Driving Scene Understanding (II-DSU) Model to obtain the driving scene understanding implicitly, as concisely shown in the lower line of ~\cref{fig:pipeline}. The main idea is to put the scene understanding task into the end-to-end model and use the implicit high-dimensional feature in the model as the scene understanding results, thus tackling the issues mentioned above. However, a new problem is how to explain and validate the implicit form of the scene understanding in the end-to-end model. In particular, we directly use the final result, planning trajectory, to verify whether the high-dimensional scene understanding results can support the complicated driving task. In addition, we add four auxiliary perception tasks to provide more useful priori scene knowledge when training the model and to get a partial observation of the understanding feature during inference. We summarize the contributions of our paper as follows:
    \begin{itemize}
    \item We propose an end-to-end Interpretable Implicit Driving Scene Understanding (II-DSU) model to obtain a high-dimensional driving scene understanding feature through the guide provided by the downstream planning.
    \item We add four explicit auxiliary perception tasks to train the model for getting a more comprehensive understanding and to make the implicit results visible.
    \item We experimentally validate the performance of our method on CARLA benchmarks with complicated urban scenarios. The results show that our model outperforms prior methods.
    \end{itemize}
    
\section{Related Work}
    \begin{figure*}[!h]
    \vspace*{5pt}
        \centering
        \includegraphics[width=13cm]{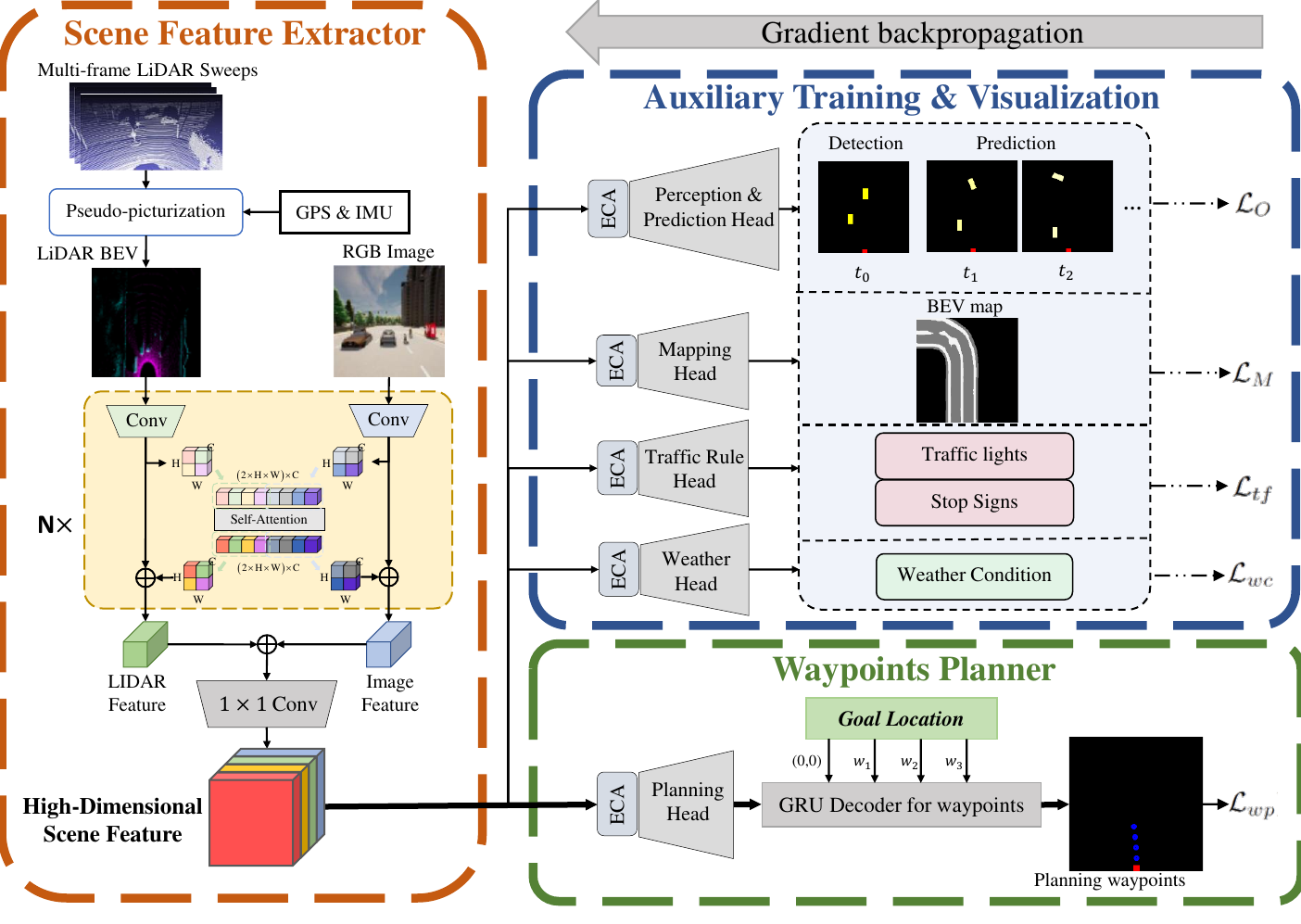}
        \caption{\textbf{Architecture of our II-DSU model.} Each LiDAR sweep is firstly converted into a 2-dimensional BEV grid map. Then all frames are aligned and stacked into a multi-channel LiDAR pseudo-image. The obtained LiDAR pseudo-image and RGB image are passed through feature encoder and transfuser layers which use the self-attention mechanism to exchange the information between the features of two branches. Finally, the features are concatenated to obtain a high-dimensional scene feature. The downstream planning module and multiple auxiliary perception tasks are used to train the model, while the scene feature can also be indirectly evaluated by the planning results and visualized by the auxiliary perception tasks.}
        \label{fig:model}
        \vspace{-2mm}
    \end{figure*}

    \noindent{\bf Explicit Understanding\textemdash Scene Graph Generation:} A scene graph is a structured representation of the driving scene which can clearly represent the instances in the scene, the attributes of the instances, and the relationships between the instances. Using scene graphs as an explicit representation of scene understanding can contain richer scene information than other object-level perception tasks.
    
    Existing scene graph generation (SGG) methods usually consist of two phases: object detection and predicate estimation. Given the detection results, early approaches used conditional random fields~\cite{dai2017detecting} to predict predicates or transformed the predicate prediction into a classification problem ~\cite{qi2019attentive}. Recently, researchers~\cite{chen2019knowledge, lin2020gps} use explicit methods of graph-based modeling and RNN to solve this question.~\cite{yang2022panoptic} propose a Panoptic Scene Graph (PSG) method based on a pixel-level panoptic segmentation, which can locate instances correctly and establish the relationship between the background and foreground instances. For self-driving tasks, ~\cite{tian2021rsg} first propose the concept of Road Scene Graph (RSG) and validate their RSG-Net on their individual dataset.~\cite{malawade2022spatiotemporal} attempt to use multi-frame image sequences to generate RSG and apply it to the downstream collision prediction task. They test their model with both real-world and virtual datasets to verify whether SGG is transferable in the self-driving field. However, this explicit understanding method cannot solve the problems mentioned before. 
    
    \noindent{\bf Implicit Understanding\textemdash Interpretable End-to-end Driving Model:} Research on end-to-end driving has become more popular and is basically divided into two categories: reinforcement learning (RL) and imitation learning (IL). Reinforcement learning usually requires learning in an interactive environment.~\cite{toromanoff2020end} first applied the online RL process to a complex urban driving scenario. It can effectively obey traffic rules to successfully complete an autonomous driving task. Imitation learning, on the other hand, uses collected expert data (waypoints or control signals) for the model to imitate. LBC~\cite{chen2020lbc} and NEAT~\cite{chitta2021neat} regress the waypoints from a dense heatmap or offset map after decoding the extracted feature. However, the complexity of these models is high because the corresponding heatmap or offset map needs to be computed for each frame. TransFuser~\cite{prakash2021multi} uses a simple GRU network directly regress the waypoints. LAV~\cite{chen2022lav} further proposes a temporal iterative GRU network to refine the waypoints.
    
    However, most of these approaches focus on how to perform reasonable planning. For end-to-end models, interpretability is another major problem to be solved.~\cite{zeng201nmp} is the first to propose an interpretable end-to-end planning model, which mainly uses point clouds and HD maps as inputs and generates intermediate visible detection and prediction results. To avoid depending on HD maps excessively,~\cite{casas2021mp3} proposes MP3 using raw data to generate an online map, detection, and prediction results, and then put them into a sampling-based trajectory planner. Recently, InterFuser~\cite{shao2022safety} tries to design auxiliary tasks to provide more interpretable feature results through tokens in the attention mechanism. Yet, none of these approaches considers getting a whole scene understanding feature and making it interpretable. In contrast, we obtain just one common feature representing scene understanding which can be easily utilized for downstream tasks and use planning results as well as several perception tasks to interpret it.

\section{Method}
    In this work, we propose a method for obtaining and validating implicit driving scene understanding (as shown in~\cref{fig:model}) with three aspects: (1) the downstream planning task in the end-to-end model is used to guide the multimodal fusion module to get more helpful scene information for self-driving tasks; (2) four additional auxiliary perception tasks, which provide more prior scene knowledge to the model, are used to assist model’s training; (3) the planning trajectory is used to {\em indirectly} reflect the goodness of implicit scene understanding, and the results of auxiliary perception tasks partly visualize the implicit scene understanding to {\em directly} reflect whether the scene understanding feature is sufficient to support downstream tasks.

\subsection{Input and Output Representations}

    \noindent{\bf Input representations:} For more general scene understanding results, the introduction of the time dimension is crucial. Therefore, we use LiDAR data, which is more representative of the change in scene 3D information, as multi-frame input, denoted as  $\left\{{\rm lidar}_{t} \right\}_{t=-T_{past}+1}^0$. We use $T_{past}=3$, which means using two past frames and one current frame as input. To make full use of the multi-frame data and to simplify the form of input, we first align data by using GPS/IMU to transfer all the LiDAR in past frames to the coordinates where the current frame’s ego vehicle is taken as the origin. Then, following the example of~\cite{rhinehart2019precog}, each frame of the LiDAR point cloud is converted into a 2-bin histogram over a 2-dimensional Bird’s Eye View (BEV) grid $\left\{{\rm BEV\_lidar}_{t} \in \mathbb{R}^{R \times R \times 2} \right\}_{t=1-T_{past}}^0$. The area considered by the 2D BEV grid map is with 32 m distance in front of the vehicle and 16 m to each of the sides. The ratio is 8 pixels per meter which results in a resolution of 256$\times$256 pixels. For the histogram, the 2 bins present the points above and at ground level. Intuitively, the channel at ground level usually reflects the static scene information, while the one above the ground level contains movable objects such as cars and pedestrians. So, for multi-frame LiDAR, we directly add the “at ground level” channels into one channel to enhance the static feature of ground and curb, and concatenate the “above ground level” channels to retain the dynamic information, resulting in a ($T_{past}+1$)-channel LiDAR BEV image input ${\rm {\bf I}_{lidar}}$ of shape ${256\times256\times4}$. For the RGB input, one front camera with a 100$^{\circ}$ FOV and a 400$\times$300 resolution in pixels is used and we crop the images as image input ${\rm {\bf I}_ {image}}$ of shape ${256\times256\times3}$.
    
    \noindent{\bf Output representations:} For driving scene understanding task, we finally get an implicit general scene understanding feature ${\rm \bf F}_{DSU}$ with the shape of 8$\times$8$\times$1024. However, in order to train and validate the whole end-to-end model, the outputs of the downstream planning task and the four auxiliary perception tasks are required to calculate the loss and to visualize. As in~\cite{prakash2021multi}, the planning task outputs the future trajectory {\bf w} of the ego vehicle in BEV space, which coincides with the ego vehicle’s current coordinates frame, to guide the vehicle to steer toward. The trajectory is represented by a series of 2D waypoints $\left\{{\rm w}_{t}= \left(x_t,y_t \right) \right\}_{t=1}^{T_{future}}$, where $T_{future}=4$. Then two PID controllers are connected to transform the waypoints into the control signals. For the four perception tasks in the auxiliary training and visualization module, the outputs consist of object density map, BEV map, traffic rule information, and weather condition. The object density map $O \in \mathbb{R}^{R \times R \times \left(7 \times 3 \right)}$ is used for detecting and predicting the agents in the next three timesteps. Each agent can be parameterized as $(x, y,  w, l, \theta)$ from the BEV, corresponding to the center position, size, and orientation. Therefore, for each timestep, the provided 7 features represent a 1-channel heatmap and a 6-channel regression map $(dx, dy, \log(w), \log(l), sin, cos)$, where $(dx, dy)$ denotes the pixel offset of the object center, and $sin/cos$ denotes the $\tan(\theta)$. The Bev map $M \in \mathbb{R}^{R \times R \times 3}$ predicts the BEV segmentation map of the drivable area, lines and other. The traffic rule information contains the state of the traffic lights and whether there is a stop sign in front of the ego vehicle. Besides, the model also classifies the weather condition into sunny, cloudy, rainy, and foggy which may influence the agents’ driving behavior.

\subsection{Model Architecture Design}

\begin{figure}[t]
    \vspace*{5pt}
        \centering
        \includegraphics[clip, width=8.5cm]{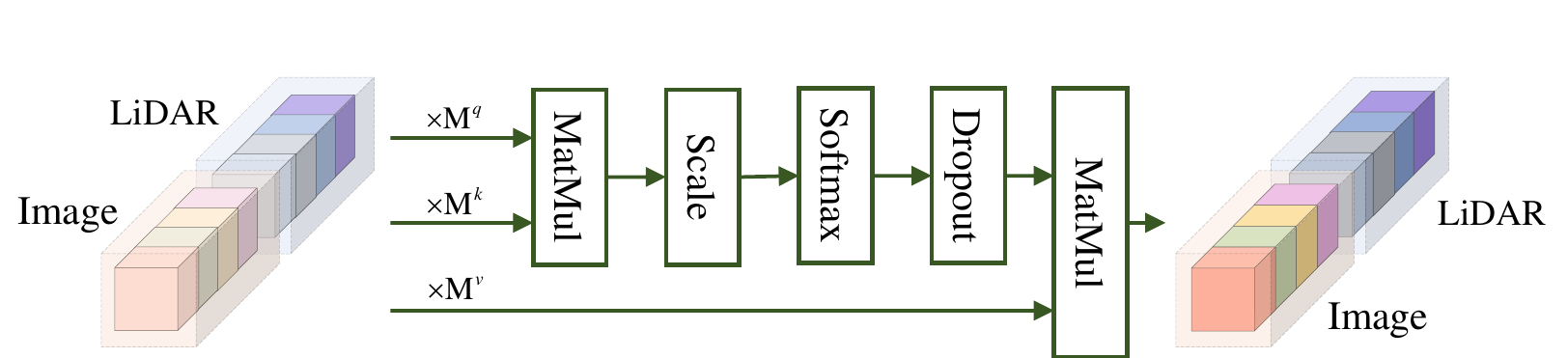}
        \caption{Detailed illustration of the transfuser layer.}
    \label{fig:self-attention}
    \vspace{-2mm}
    \end{figure}

    \noindent{\bf Scene Feature Extractor:} To extract a general driving scene feature from ${\rm {\bf I}_{lidar}}$ and ${\rm {\bf I}_ {image}}$, four transfuser layers~\cite{prakash2021multi} (\cref{fig:self-attention}) are utilized to fuse LiDAR and image modalities given their complementary nature. The ${\rm {\bf I}_{lidar}}$ and ${\rm {\bf I}_ {image}}$ first go through ConvNeXt-Tiny~\cite{Liu_2022_CVPR} separately to get different dimension features with the convolutional channels of (64, 128, 256, 512) and the corresponding down-sampling factors are (/4, /2, /2, /2). For each dimension, the features are split into equal-sized patches. Then all patches are used as tokens and added the position encoding. Each feature patch is updated using the self-attention mechanism and then sent back to the place from which they are extracted before fusion. Finally, the highest dimension of the image and LiDAR features are concatenated and put into a 1$\times$1 convolution network to get ${\rm \bf F}_{DSU}$.

    \noindent{\bf ECA Module:} Before ${\rm \bf F}_{DSU}$ is sent to each head, we add an efficient channel attention (ECA) module~\cite{wang2020eca}, and each head's ECA can add weights of channels to the high-dimensional feature that this head's task depends on. Benefiting from just one high-dimensional feature presenting the scene, we can calculate how each task helps the model to learn scene information by using the cosine similarity.
    
    \noindent{\bf Planning Header:} The main output of the end-to-end model is the downstream planning task. We design the planning header as~\cite{prakash2021multi}. The channel-weighted implicit high-dimensional feature is reduced to a dimension of ${1\times1\times1024}$ by  average pooling and flattened to a 1024-dimensional vector. The 1024-dimensional vector is then reduced to 128 dimensions by a 3-layer MLP. Finally, the reduced features, the differential ego vehicle waypoints, and the relative goal points are fed into four cascaded Gated Recurrent Units (GRUs)~\cite{cho2014GRU} to obtain the final waypoint {\bf w}.
    
    \noindent{\bf Auxiliary Perception Headers} For the four perception tasks in the auxiliary training and visualization module, the channel-weighted scene feature is followed by four parallel perception headers to predict the object density map $O$, the BEV map $M$, the traffic rule, and the weather condition. For the object density map, the feature is directly upsampled with a CenterNet Decoder~\cite{zhou2019objects} and then sent to two $1 \times1$ convolution networks to predict the heatmap and the regression map. Similarly, a convolutional decoder is also used to obtain the BEV map $M$. For traffic rule and weather condition prediction, the feature is reduced to 1024 dimensions. Then a fully connected layer and an activate function are used to calculate the classification score. Specifically, the sigmoid and softmax function are utilized for predicting the traffic rule information and weather condition respectively.
    
\subsection{Loss Function}
    The training of II-DSU is a multi-task learning process due to the introduction of auxiliary perception tasks, so the loss function is designed with the following components to train the scene feature extractor and all headers: the predicted waypoints ($\mathcal{L}_{wp}$), the object density map ($\mathcal{L}_{O}$), the BEV map ($\mathcal{L}_{M}$), the traffic rule information ($\mathcal{L}_{tf}$) and the weather condition ($\mathcal{L}_{wc}$):
    \begin{equation}
        \mathcal{L} = {\lambda_{wp}}\mathcal{L}_{wp} + {\lambda_{O}}\mathcal{L}_{O} + {\lambda_{M}}\mathcal{L}_{M} + \\ {\lambda_{tf}}\mathcal{L}_{tf} + {\lambda_{wc}}\mathcal{L}_{wc},
        \label{total_loss}
    \end{equation}
    \noindent where $\lambda$ are hyperparameters for balance and we expect the planning task to dominate the training. 
    
    The waypoints regression loss $\mathcal{L}_{wp}$ is set as the L1 loss between the expert’s and the agent’s waypoints. For the $\mathcal{L}_{M}$ and $\mathcal{L}_{wc}$ loss, we both use the cross-entropy loss to classify.
    The object density map loss $\mathcal{L}_{O}$ consists of two parts: 
    \begin{equation}
        \mathcal{L}_{O} = \alpha \mathcal{L}_{heat} + \beta \mathcal{L}_{reg},
        \label{O_loss}
    \end{equation}
    \noindent where $\mathcal{L}_{heat}$ is a cross-entropy loss for heatmap, and $\mathcal{L}_{reg}$ is a smooth L1 loss for box parameters regression. 
    
    When predicting the traffic information $\mathcal{L}_{tf}$, we expect to recognize the traffic light status $\mathcal{L}_{light}$ and stop sign $\mathcal{L}_{sign}$:
    \begin{equation}
        \mathcal{L}_{tf} = \gamma \mathcal{L}_{light} + \delta \mathcal{L}_{sign},
        \label{_loss}
    \end{equation}
    \noindent which are calculated by binary cross-entropy loss and $\alpha, \beta, \gamma, \delta$ are used to balance the loss items.

    \begin{table*}
    \vspace*{5pt}
        \centering
        \renewcommand\arraystretch{1.05}
        \begin{subtable}[t]{0.55\linewidth}
        \setlength\tabcolsep{3pt}
        \resizebox{\linewidth}{!}{
            \begin{tabular}{c|cc|cc}
                \toprule[0.5mm]
                \multirow{2}{*}{\textbf{Method}} & \multicolumn{2}{c}{\textbf{Town05 Short}}\vline & \multicolumn{2}{c}{\textbf{Town05 Long}} \\ 
                \cline{2-5}
                 & DS$\uparrow$ & RC$\uparrow$ & DS$\uparrow$ & RC$\uparrow$ \\
                \midrule[0.5mm]
                CILRS~\cite{codevilla2019exploring} & $ 7.47\pm2.51 $ & $ 13.40\pm1.09 $ & $ 3.68\pm2.16 $ & $ 7.19\pm2.95 $ \\
                LBC~\cite{chen2020lbc} & $30.97\pm4.17$ & $55.01\pm5.14$ & $7.05\pm2.13$ & $32.08\pm7.40$ \\
                AIM~\cite{prakash2021multi} & $49.00\pm6.83$ & $81.07\pm15.59$ & $26.50\pm4.82$ & $60.66\pm7.66$ \\
                TransFuser~\cite{prakash2021multi} & $54.52\pm4.29$ & $78.41\pm3.75$ & $33.15\pm4.04$ & $56.36\pm7.14$ \\
                NEAT~\cite{chitta2021neat} & $58.70\pm4.11$ & $77.32\pm4.91$ & $37.72\pm3.55$ & $62.13\pm4.66$ \\
                WOR~\cite{chen2021wor} & $64.79\pm5.53$ & $87.47\pm4.68$ & $44.80\pm3.69$ & $82.41\pm5.01$ \\
                TransFuser$^{*}$~\cite{chitta2022transfuser} & $82.57\pm2.52$ & $89.59\pm1.63$ & $53.04\pm3.54$ & $84.83\pm2.43$ \\
                TCP~\cite{wu2022trajectory} & $75.74\pm3.24$ & $88.44\pm1.50$ & $53.27\pm4.21$ & $85.34\pm2.14$ \\
                InterFuser$^\diamondsuit$ & $82.60\pm4.24$ & $92.53\pm3.44$ & $58.28\pm5.64$ & $88.12\pm3.47$ \\
                \textbf{II-DSU (Ours)} & $\textbf{87.73}\pm2.93$ & $\textbf{94.09}\pm2.93$ & $\textbf{60.28}\pm4.03$ & $\textbf{91.88}\pm2.60$ \\
                \hline
                \textit{InterFuser$^\spadesuit$}~\cite{shao2022safety} & \textit{94.95} $\pm$ \textit{1.91} & \textit{95.19} $\pm$ \textit{2.57} & \textit{68.31} $\pm$ \textit{1.86} & \textit{94.97} $\pm$ \textit{2.87} \\
                \bottomrule[0.5mm]
            \end{tabular}}
            \caption{Comparison of our II-DSU with other methods in Town05 benchmark composed with Town05 Short and Town05 Long settings on 2 metrics: DS and RC.}
            \label{tab:result_a}
        \end{subtable}
        \begin{subtable}[t]{0.42\linewidth}
        \setlength\tabcolsep{3pt}
        \resizebox{\linewidth}{!}{
            \begin{tabular}{c|ccc}
                \toprule[0.5mm]
                \textbf{Method} & \textbf{DS}$\uparrow$ & \textbf{RC}$\uparrow$ & \textbf{IS}$\uparrow$ \\
                \midrule[0.5mm]
                CILRS~\cite{codevilla2019exploring} & $ 22.97\pm0.90 $ & $ 35.46\pm0.41 $ & $ 0.66\pm0.02 $ \\
                LBC~\cite{chen2020lbc} & $29.07\pm0.67$ & $61.35\pm2.26$ & $0.57\pm0.02$ \\
                AIM~\cite{prakash2021multi} & $51.25\pm0.17$ & $70.04\pm2.31$ & $0.73\pm0.03$ \\
                TransFuser~\cite{prakash2021multi} & $53.40\pm4.54$ & $72.18\pm4.17$ & $0.74\pm0.04$ \\
                NEAT~\cite{chitta2021neat} & $65.10\pm1.75$ & $79.17\pm3.25$ & $0.82\pm0.01$ \\
                WOR~\cite{chen2021wor} & $67.64\pm1.26$ & $90.16\pm3.81$ & $0.75\pm0.02$ \\
                TransFuser$^{*}$~\cite{chitta2022transfuser} & $71.92\pm2.84$ & $83.24\pm3.68$ & $0.84\pm0.02$ \\
                TCP~\cite{wu2022trajectory} & $74.60\pm3.01$ & $80.64\pm3.11$ & $\textbf{0.93}\pm0.01$ \\
                InterFuser$^\diamondsuit$ & $79.46\pm2.92$ & $90.24\pm4.75$ & $0.86\pm0.02$ \\
                \textbf{II-DSU (Ours)}& $\textbf{86.79}\pm2.08$ & $\textbf{94.93}\pm2.25$ & $0.91\pm0.01$ \\
                \hline
                \textit{InterFuser$^\spadesuit$}~\cite{shao2022safety} & \textit{91.84} $\pm$ \textit{2.17} & \textit{97.12} $\pm$ \textit{1.95} & \textit{0.95} $\pm$ \textit{0.02} \\
                \bottomrule[0.5mm]
            \end{tabular}}
            \caption{Comparison of our II-DSU with other methods in CARLA 42 Routes benchmark on 3 metrics: DS, RC and IS.}
            \label{tab:result_b}
        \end{subtable}
        \caption{\textbf{Performance results.} We compare our II-DSU model with CILRS, LBC, AIM, TranFuser, NEAT, WOR, TransFuser$^*$, TCP, and InterFuser in both of Town05 benchmark (\cref{tab:result_a}) and CARLA 42 Routes benchmark (\cref{tab:result_b}). Metrics: driving score(DS), route completion(RC), and infraction score(IS). Our method performed better than all other strong methods in all metrics and scenarios. The $\spadesuit$/$\diamondsuit$ indicates InterFuser with/without a safety controller which is sophisticated hand-designed post-processing based on the intermediate outputs.}
        \label{tab:result}
    \end{table*}

    \begin{table*}
        \centering
        \renewcommand\arraystretch{1.05}
        \begin{tabular}{c|ccc|cccccccc}
            \toprule[0.5mm]
            & \textbf{DS}$\uparrow$ & \textbf{RC}$\uparrow$ & \textbf{IS}$\uparrow$ & \textbf{Ped}$\downarrow$ & \textbf{Veh}$\downarrow$ & \textbf{Lay}$\downarrow$ & \textbf{Red}$\downarrow$ & \textbf{OR}$\downarrow$ & \textbf{Dev}$\downarrow$ & \textbf{TO}$\downarrow$ & \textbf{Block}$\downarrow$  \\ 
            \midrule[0.5mm]
            \textit{None} & 59.28 & 75.15 & 0.81 & 0.19 & 1.11 & 0.21 & 0.14 & 1.90 & 0.02 & 1.91 & 3.38 \\
            \hline
            No Multi-frame LiDAR & 84.72 & 93.90 & 0.89 & 0.14 & 0.72 & \textbf{0.09} & 0.09 & 1.39 & \textbf{0.00} & 1.34 & 2.87 \\
            \hline
            No Detection \& Prediction & 75.93 & 88.94 & {\color {black} {0.88}} & {\color {blue} {0.18}} & {\color {blue} {1.06}} & {\color {black} {0.14}} & 0.09 & 1.37 & {\color {black} {\textbf{0.00}}} & {\color {black} {1.36}} & 3.07\\
            No BEV Map & {\color {blue} {74.61}} & {\color {blue} {88.11}} & 0.87 & 0.12 & 0.69 & {\color {blue} {0.21}} & {\color {black} {0.06}} & {\color {blue} {1.84}} & {\color {blue} {0.01}} & 1.51 & {\color {black} {2.54}}\\
            No Traffic Rule & {\color {black} {81.51}} & {\color {black} {94.12}} & {\color {black} {0.88}} & {\color {black} {0.11}} & {\color {black} {0.63}} & 0.16 & {\color {blue} {0.16}} & {\color {black} {1.37}} & {\color {black} {\textbf{0.00}}} & 1.39 & {\color {blue} {3.14}}\\
            No Weather & 78.83 & 93.36 & {\color {blue} {0.86}} & {\color {blue} {0.18}} & 0.97 & {\color {black} {0.14}} & {\color {black} {\textbf{0.08}}} & 1.42 & {\color {black} {\textbf{0.00}}} & {\color {blue} {2.00}} & 2.68\\
            \hline
            {All tasks} & {\textbf{86.79}} & {\textbf{94.93}} & {\textbf{0.91}} & {\textbf{0.10}} & {\textbf{0.56}} & {0.11} & {\textbf{0.08}} & {\textbf{1.32}} & {\textbf{0.00}} & {\textbf{1.21}} & {\textbf{2.44}}\\
            \bottomrule[0.5mm]
        \end{tabular}
            
        \caption{\textbf{Ablation study for multi-frame input and auxiliary tasks.} The results are evaluated on the CARLA 42 Routes benchmark. Besides the 3 main metrics, we also introduce more infraction metrics used to calculate IS in detail(Ped: Collisions with pedestrians, Veh: Collisions with vehicles, Lay: Collisions with the scene layout, Red: Red light infractions, OR: Off-road infractions, Dev: Route deviation, TO: Route timeouts, Block: Agent blocked). The data marked in {\color {blue} {blue}} are the worst in each column among the middle four rows.}
        \label{tab:ablation}
        \vspace{-2mm}
    \end{table*}
    
\section{Experiments}

\subsection{Experiment Setup}

    In order to validate our model’s ability to understand scenes, we primarily use the metrics in the downstream planning task to reflect whether the high-dimensional scene understanding feature is sufficient to support the overall self-driving task. Thus, as with most end-to-end driving models, we implement our model on the open-source CARLA~\cite{dosovitskiy2017carla} simulator with version 0.9.10.1.
    
    \noindent{\bf Dataset Collection:} We run a simple handcrafted rule-based expert agent similar to~\cite{prakash2021multi} on 8 kinds of towns in CARLA with 2 FPS, to collect an expert training dataset of 146k frames in total. For providing abundant diverse scene information, we randomly set different routes and spawn variable agents. Notice that we directly reflect the effects of weather conditions in data. For instance, we randomly set the driving behavior of all traffic agents to \emph{Normal} or \emph{Aggressive} on sunny and cloudy days and to \emph{Cautious} on rainy or foggy days. These behavior classes include the speed limit, the safety time, the minimum safety distance, and other safety-related parameters.
    
    \noindent{\bf Benchmark:} We evaluate our methods on the Town05 benchmark~\cite{prakash2021multi} and CARLA 42 Routes benchmark~\cite{chitta2021neat}. The Town05 benchmark considers two evaluation settings: (1) Town05 Short: 10 short routes of 100-500m comprising 3 intersections each, (2) Town05 Long: 10 long routes of 1000-2000m comprising 10 intersections each. The CARLA 42 Routes benchmark consists of 42 routes from Town01 to Town06 with different weather and daylight conditions. Both of the benchmarks have dynamic agents and complicated situations which make the self-driving task more challenging.
    
    \noindent{\bf Metrics:} As we mentioned before, we use the results of the planning task to verify whether the model extracts a general scene information feature sufficient to accomplish the self-driving task. So, in this section, we explain the unified metrics introduced by the CARLA Leaderboard for evaluating the driving task: the \textbf{route completion} ratio (\textbf{RC}), the \textbf{infraction score} (\textbf{IS}), and the \textbf{driving score} (\textbf{DS}). RC shows the percentage of the completed route. IS is a cumulative multiplicative penalty for each violation to show the route’s overall performance discount. DS is averaged by multiplying the RC and IS for each route, which gives a more comprehensive assessment of the driving.

\subsection{Results}
    
    \cref{tab:result} compare our method with several end-to-end driving models from the CARLA Leaderboard on the Town05 benchmark and CARLA 42 Route benchmark separately. It is noteworthy that InterFuser$^\spadesuit$~\cite{shao2022safety} uses complex post-processing for trajectory modification and control signal generation to achieve good results. It is unfair to compare it to other models, especially those trajectory-based methods using PID controllers. Therefore, we replace the safety controller with a regular PID controller as InterFuser$^\diamondsuit$. It should also be mentioned that our II-DSU model has the ability to use intermediate results to design a similar safety controller for better driving performance as what InterFuser$^\spadesuit$ does.

    A row-wise data comparison in~\cref{tab:result_a} and~\cref{tab:result_b} directly shows that our model achieves a significant performance improvement in self-driving tasks. In the Town05 Short (Long) setting, the DS and RC increase by about 6.2\% (3.4\%) and 1.7\% (4.3\%) respectively. In the CARLA 42 Routes benchmark, the DS and RC also get 9.2\% and 5.2\% boosts respectively. We find that the improvement of DS and RC in CARLA 42 Route benchmark is higher than the one in Town05 benchmark, which indicates that our model can handle more complex scenarios and the weather prediction task can obtain more priors from the CARLA 42 Routes benchmark with various weather conditions. 
    
    Furthermore, TransFuser~\cite{prakash2021multi} is the best baseline for our model, as we use a similar feature fusion mechanism and a downstream planning GRU module. But for TransFuser, just a one-dimensional vector, which presents the whole scene, is used for the downstream module to plan. In contrast, we use a high-dimensional feature representation of the scene information. Moreover, throughout the learning process of high-dimensional scenes, there are more effective auxiliary perception tasks strongly related to self-driving than casually selecting some visual tasks. These tasks can provide more useful prior knowledge of the scenes for the models to learn, which is finally reflected in better driving performance compared to the TransFuser$^*$~\cite{chitta2022transfuser}. All of the above means that II-DSU can obtain richer and more helpful understanding results for the self-driving task, given the similar structure of the downstream planning model in the baseline.

    \begin{table}[t]
    \vspace*{5pt}
        \centering
        \renewcommand\arraystretch{1.05}
        \setlength{\tabcolsep}{8mm}
        \begin{tabular}{c|c}
            \toprule[0.5mm]
            & cosine similarity \\ 
            \midrule[0.5mm]
            plan $\times$ det\&pred & 0.6465\\
            plan $\times$ BEV & 0.6154\\
            plan $\times$ traffic & 0.2177\\
            plan $\times$ weather & 0.3253\\
            \bottomrule[0.5mm]
        \end{tabular}   
        \caption{\textbf{Correlation of auxiliary tasks and planning.}}
        \label{tab:cosine similarity}
        \vspace{-2mm}
    \end{table}

\subsection{Ablation Study}
    
    Since the scenario of the Town05 benchmark does not introduce diverse weather and thus cannot reflect the role of the weather header proposed in our model, we conduct ablation experiments on the CARLA 42 Routes benchmark to investigate how each of the four auxiliary perception tasks affects scene understanding and still validate the results with downstream planning. In addition to the three metrics used previously, we also introduce specific penalty metrics, which are used to calculate the IS, for a comprehensive analysis.
    
    In~\cref{tab:ablation}, we evaluate the impact of multi-frame LiDAR input and four auxiliary perception tasks on driving performance (scene understanding) within a baseline (\textit{None}) similar to TransFuser. As can be seen from the overall results, adding multi-frame LiDAR and all auxiliary perception tasks for training results in the best driving performance in primary metrics. 
    
    Furthermore, we need to focus on the data in the middle fourth to seventh rows to analyze the role of each auxiliary task. \textbf{No Detection \& Prediction.} \textit{Collisions with pedestrian} and \textit{Collisions with vehicles} both become significantly higher (lower is better), indicating that the model does not effectively learn and pay attention to the information of traffic participants in the scenario, which leads to the collisions. \textbf{No BEV Map.} When map information is not available, there is a significant decrease in RC and DS, and the worst results are obtained for \textit{Collisions with the scene layout}, \textit{Off-road infractions} and \textit{Route Deviations}, which means that the hard constraints such as roads or lanes information are critical to the self-driving task. \textbf{No Traffic Rule.} The worst result for \textit{Red Light Infractions} is obvious when the model pays less attention to the traffic lights and stop signs. Then the ego-vehicle is more likely to get stuck at crossroads by other vehicles. \textbf{No Weather.} The model behaves more conservatively in bad weather conditions because we deliberately set different agent behavior for different weather scenarios during the data collection phase. When the model does not focus on the weather directly, it cannot choose the appropriate driving strategy based on the weather condition. Due to the distribution of the training data, the ego-vehicle tends to drive at a slower speed and takes more time to complete the route which results in the deterioration of \textit{Route Tiemouts} and \textit{Route timeouts}.

    Besides directly removing tasks for ablation experiments, since we only use one high-dimensional feature to represent the scene information and introduce the ECA module, we can analyze the correlation between the manual-introduced auxiliary tasks and the feature which model considers to be useful to the self-driving by calculating the cosine similarity of each auxiliary task's channel weights and the planning head's channel weights. The results are shown in~\cref{tab:cosine similarity}. This means that the information learned by the model itself is closely related to the detection \& prediction task and the BEV task. In other words, it indicates that the scene information the model focuses on contains what needs to be learned in these auxiliary tasks, and that it also contains useful scene information that cannot be represented explicitly.
    

\subsection{Visualization}
    \begin{figure}
      \vspace*{5pt}
      \centering
      \begin{subfigure}{0.3\linewidth}
        \centering
        \includegraphics[clip, width=2.57cm]{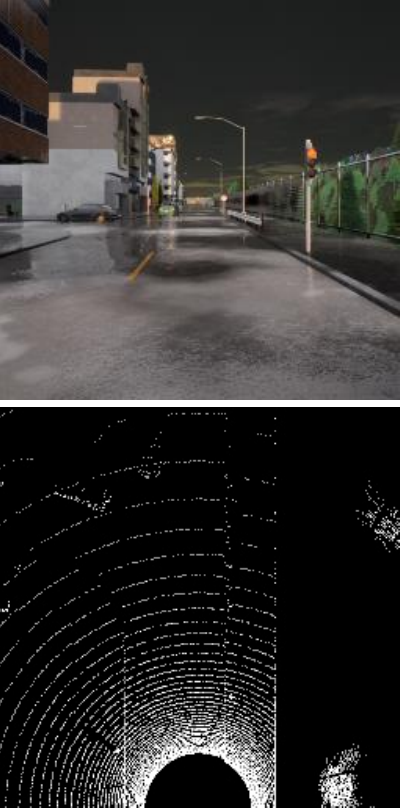}
        \caption{Top to bottom: image input and current frame LiDAR input.}
        \label{fig:visualization-a}
      \end{subfigure}
      \hfill
      \begin{subfigure}{0.64\linewidth}
        \centering
        \includegraphics[clip, width=5.2cm]{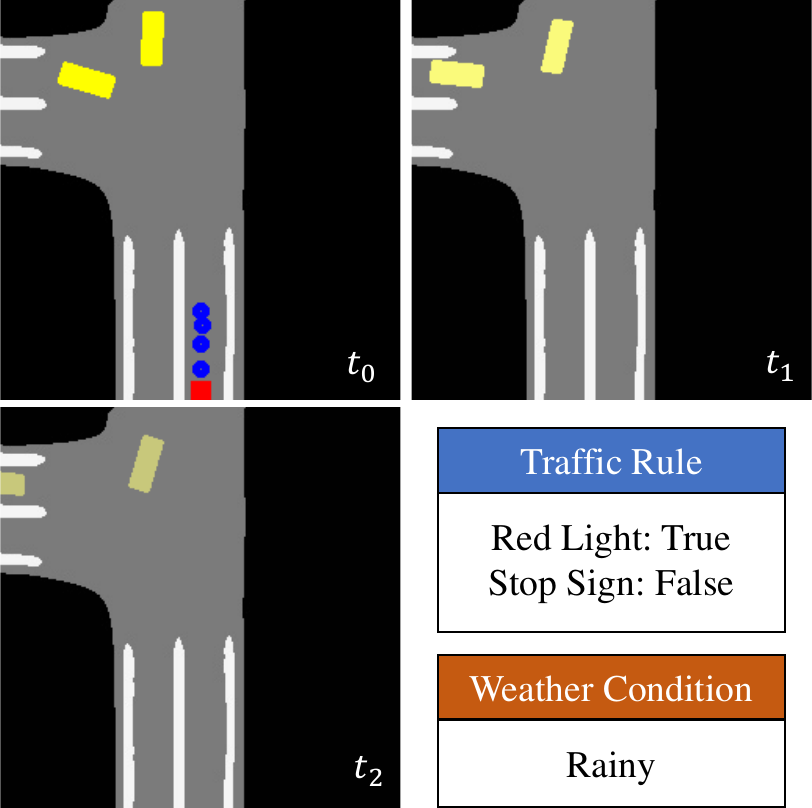}
        \caption{Top-left: predicted BEV map, object detection and ego waypoints; top-right and bottom-left: prediction; bottom-right: traffic rule and weather condition.}
        \label{fig:visualization-b}
      \end{subfigure}
      \caption{\textbf{Visualization of Auxiliary tasks.} The input and output are visualized. Different colors mean: \textcolor[RGB]{123,123,123}{drivable area (grey)}, \textcolor[RGB]{228,228,228}{lanes (white)}, \textcolor[RGB]{0,0,0}{background (black)}, \textcolor[RGB]{255,0,0}{ego-vehicle (red)}, \textcolor[RGB]{223,218,8}{other vehicle (yellow)}, \textcolor[RGB]{0,0,255}{ego waypoints (blue)}}
      \label{fig:visualization}
      \vspace{-2mm}
    \end{figure}
    
    \begin{figure}[t]
        \centering
        \includegraphics[clip, width=\linewidth]{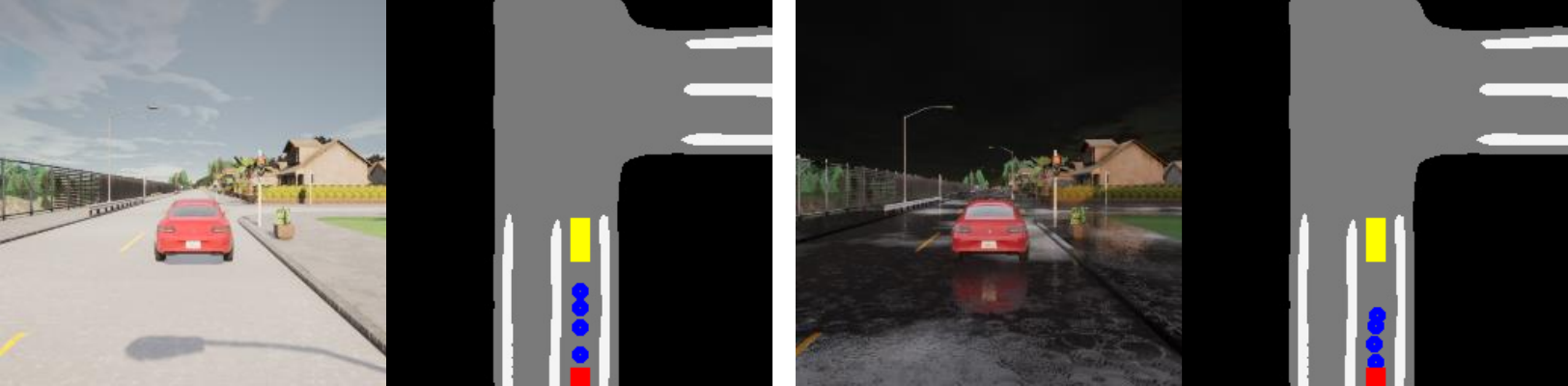}
        \caption{\textbf{Comparison of the planning waypoints under different weather conditions.} The first row shows a sunny day. The second row keeps the same scene and only changes the weather to rainy.
        }
        \label{fig:comparison}
        \vspace{-2mm}
    \end{figure}
    
    In~\cref{fig:visualization}, we visualize the ego waypoints predicted by our models and all outputs of four auxiliary perception tasks. We can find that the model reasonably outputs the planning waypoints of the ego-vehicle and gets intuitive results of auxiliary perception tasks, which indicates that the high-dimensional scene feature extracted by our model contains rich scene information to support these downstream tasks.

    In particular, we also visualize and compare the model output for the same scenario under different weather conditions as shown in~\cref{fig:comparison}. It can be seen that II-DSU outputs more conservative planning waypoints on a rainy day, thus keeping the distance from the front car. This is similar to the real driving situation and ensures driving safety.



\section{Conclusion}
\label{sec:conclusion}  
   
   Since the scene understanding task cannot be replaced by specific perception tasks, we propose II-DSU that uses multi-frame multimodal data for driving scene understanding and ultimately obtains a general high-dimensional feature to represent the scene. The process of comprehension in this model is guided by a downstream planning task and several auxiliary tasks so that it can focus on more driving-related information. The ECA module allows us to calculate the extent to which the auxiliary tasks assist the planning task. In addition, the implicit feature is difficult to interpret, so the results of the downstream planning provide an indirect verification of scene understanding, and four auxiliary perception tasks also provide visualization. The future direction of this work could be how to effectively demonstrate the generality of the understanding feature we obtained, i.e., whether it can directly replace the model’s backbone of other perception tasks and accomplish that task with only slight fine-tuning.

{\small
\bibliographystyle{ieee_fullname}
\bibliography{egbib}
}

\end{document}